\documentclass{article}
\usepackage{iclr2027_conference,times}
\usepackage{amsmath,amssymb}
\usepackage{graphicx,booktabs,multirow,array,tabularx}
\usepackage{tikz}
\usetikzlibrary{arrows.meta,calc}
\usepackage{hyperref}
\usepackage{url}
\usepackage{placeins}
\newcommand{\AC}{Allen\textendash Cahn}
\newcommand{\GS}{Gray\textendash Scott}
\newcommand{\MZ}{Mori\textendash Zwanzig}
\hypersetup{hidelinks,pdftitle={Neural Fields Encode Adaptation Geometry},pdfauthor={Prateik Sinha and Stefania Druga}}

\title{
Neural Fields Encode Adaptation Geometry
}
\author{
Prateik Sinha\thanks{Work done during an internship at Sakana AI.} \\
Machine Learning Department \\
Carnegie Mellon University \\
Pittsburgh, PA 15213, USA \\
\texttt{prateiks@cs.cmu.edu} \\
\And
Stefania Druga \\
Sakana AI \\
Tokyo, Japan \\
\texttt{stefania@sakana.ai}
}
\iclrfinalcopy

\begin{document}
\maketitle
\lhead{Preprint. Under review}
\begin{abstract}
Neural fields are usually evaluated by how well they reconstruct an observation. We show that this misses two useful properties of a fitted network: how easily it can adapt to new observations, and what its weights retain from earlier ones. We study these properties as \emph{adaptation geometry}.
For images, we meta-learn class-specific initializations, adapt each one to a new image, and measure how much the network must change to fit it. A simple local linear model closely predicts this adaptation cost, while replacing one network's tangent kernel with another's substantially worsens the prediction. 
Adaptation thus depends on the local geometry of the fitted network, not only on its current reconstruction.
For physical fields, we repeatedly fit the same network to observations from a sequence. Its weights then retain information about that history. When two wave histories end at exactly the same observation, the final weights recover the sign of the wave velocity with 68.6\% accuracy, whereas the current observation alone contains no such information and gives 50\%.
These two phenomena are quantitatively linked: tangent-kernel eigenvalues predict both which changes are easy to learn and how quickly they are overwritten by later fitting. 
Together, these results show that neural fields contain useful information beyond what they currently reconstruct: in how they can change and in how they got there.
\end{abstract}

\section{Introduction}
\label{sec:introduction}

Suppose we fit a small neural network to each frame of a physical simulation. We could start every fit from the same initialization, or continue training the network fitted to the previous frame. Both procedures produce a network that reconstructs the current frame. But are the resulting networks equally useful representations?
This question is natural for \emph{neural fields}, also called implicit neural representations (INRs): networks that map coordinates to signal values, such as pixel positions to image intensities \citep{sitzmann2020siren}. Their fitted weights can be used as downstream data \citep{dupont2022functa,deluigi2023inr2vec,navon2023dwsnet}, and the fitting procedure affects how useful those weights are \citep{papa2024neuralfields}. We ask what reconstruction quality alone misses.

\begin{figure}[t]
    \centering
    \includegraphics[width=0.97\linewidth]{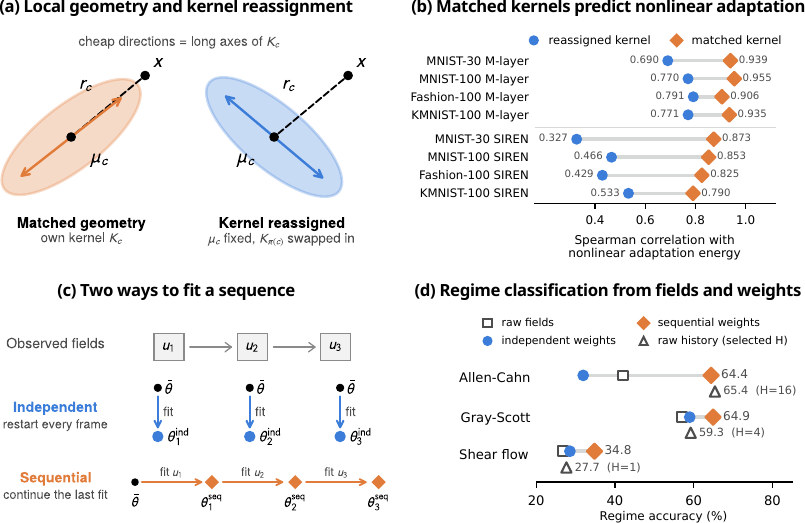}
    \caption{\textbf{Adaptation geometry and sequential fitting carry information beyond reconstruction.}
    \textbf{(a)} A class prior is its reconstruction $\mu_c$ and tangent kernel $K_c$ that sets which output changes are cheap; reassignment keeps $\mu_c$ and swaps $K_c$.
    \textbf{(b)} In all settings, the matched kernel predicts nonlinear adaptation far better than a reassigned one (Spearman correlation with nonlinear adaptation energy).
    \textbf{(c)} Independent encoding restarts every frame from the shared prior $\bar{\theta}$; sequential encoding starts each fit from the previous frame's weights.
    \textbf{(d)} With the same class-conditioned readout, sequential weights classify the regime better than raw fields or independent weights on \AC{}, \GS{}, and shear-flow (seed means); raw frames with explicit delays match them on \AC{} but not on \GS{}. The shear flow delay search selects current frames alone (Table~\ref{tab:seq}).}
    \label{fig:overview}
\end{figure}

We support this view with two main empirical results. Our contribution is not a new way to fit neural fields, but to isolate two properties of a fitted network that its current reconstruction does not reveal. First, we separate what can be predicted from the network's current output from what is gained by knowing how the network responds to further parameter updates. Second, we test whether the weights can retain information about how they were reached even when the current observation is held exactly fixed (Figure~\ref{fig:overview}).

\paragraph{Local geometry predicts adaptation.} 
In our image experiments, we learn separate initializations for each class and ask which one can be adapted most easily to a new image. A local linear model of how each network responds to small weight changes predicts the relative cost of these nonlinear adaptations and which initialization is preferred. Swapping this local geometry between classes consistently worsens those predictions, even after controlling for its overall scale
(Section~\ref{sec:static}).

\textbf{Sequential fitting creates useful memory.}
Why should history help when the underlying physical system is Markovian? A single observed field need not determine the relevant dynamics: state may be missing or reduced, or the dynamics themselves may be unknown.
The \MZ{} formalism makes the first case precise: the exact dynamics of a reduced set of observables contain a memory term that integrates over their own past \citep{mori1965transport,zwanzig1961memory}, and this is why explicit memory helps learned PDE surrogates under partial observation \citep{buitrago2025memory}. Past observations can therefore carry information about unresolved state that the current observation alone does not. We ask whether sequential neural-field
fitting can retain some of this information in its weights. Continuing each frame's fit from
the previous frame does so: we establish memory directly with wave histories that end at
identical observations, and show downstream value in physical-regime classification
(Sections~\ref{sec:collision} and~\ref{sec:seq-results}).

The two results are connected: the same tangent spectrum that determines which changes are easy to make also determines, locally, which changes are quickly overwritten. Sequential fitting therefore turns adaptation geometry into a set of memory timescales.

Finally, a controlled geometric experiment shows that the sampled reconstruction does not uniquely determine adaptation geometry: a network can follow a closed loop in output space and return to the same reconstruction with a different geometry (Section~\ref{sec:holonomy}).

\section{Related work}
\label{sec:related}

\paragraph{Neural fields, learned initializations, and local geometry.}
Neural fields represent signals with coordinate networks \citep{sitzmann2020siren}, and
prior work learns from fitted functions or their weights
\citep{dupont2022functa,deluigi2023inr2vec,navon2023dwsnet,zhou2023permutation}.
Meta-learning has also been used to learn neural-field initializations that adapt quickly to
new signals \citep{finn2017maml,nichol2018reptile,sitzmann2020metasdf,
tancik2021initializations,papa2024neuralfields}. Related work uses local network sensitivity
as a representation, as in Task2Vec \citep{achille2019task2vec}, while tangent distance,
NTKs, and linearized networks characterize local function changes under parameter updates
\citep{simard1993tangent,jacot2018ntk,lee2019wide,maddox2021linearized,
zhou2021metantk}. 
\citet{maddox2021linearized} use Jacobians of trained finite networks for analytic adaptation. Kernel Reboot \citep{Mallak_2026} combines tangent kernels with learned task priors for neural-field reconstruction and adaptation. Here we test empirically whether prior-specific tangent kernels predict rankings and choices of finite-budget nonlinear adaptation.

\paragraph{Sequential neural representations and memory.}
Warm-starting neural fields across time is not itself new: NIRVANA
\citep{maiya2023nirvana} initializes successive video representations from previous fitted
weights. 
\citet{continualmetalearningneuralfields} target catastrophic forgetting in continual neural fields,
while PDE methods evolve neural representations through time \citep{du2021ednn,bruna2024neuralgalerkin,chen2023crom}.
Our contribution is not sequential warm-starting itself,
but the analysis of what ordinary
reconstruction-based fitting leaves in its weights and whether that history is useful.

\paragraph{Memory in partially observed dynamics.}
Projecting a Markovian system onto an incomplete set of observables yields exact reduced dynamics with a Markovian term, a memory integral over the past of the observables, and a term driven by the unresolved initial state. This is the \MZ{} formalism \citep{mori1965transport,zwanzig1961memory}, later recast as optimal prediction \citep{chorin2000optimal} and connected to learned recurrent closures \citep{ma2018modelreduction}. \citet{buitrago2025memory} use this perspective to motivate memory for learned PDE surrogates and show that explicit memory helps precisely when the state is observed at low resolution or with noise.
Optimization history is also known to affect network
parameters and retention \citep{ash2020warm,kirkpatrick2017overcoming}. We connect these
views by asking whether sequential fitting itself provides a memory-bearing representation,
and whether the same local adaptation geometry that controls what is easy to change also
controls what is retained.

\section{Methods}
\label{sec:framework}

\label{sec:neural-fields}
Let $f_\theta(q)$ map a coordinate $q$ to an image intensity or physical-field value. We write $F(\theta)\in\mathbb{R}^m$ to refer to its flattened outputs. Fitting $F(\theta)$ to an observation $x$ stores a representation in $\theta$; evaluating the network decodes its reconstruction. A \emph{prior} here refers to a learned initialization for such a network. We learn it with Reptile \citep{nichol2018reptile}, which iteratively fits a network to random samples from the data and moves the initialization toward the resulting weights. For image experiments we learn a  prior for each class. Sequential experiments use one shared prior across labels. 

\subsection{Measuring which output changes are cheap}

\label{sec:framework-local}

We learn one initialization/prior $\theta_c$ for each image class, then adapt each network to reconstruct a new image and measure how much it must change to fit that image well. We compare this full adaptation with a local tangent-kernel approximation, and swap tangent kernels between classes to test whether the network's specific local geometry matters. For a class prior $\theta_c$, define the adaptation energy of a query $x$ as

\begin{equation}
\label{eq:energy}
E_c(x)=\min_\delta\left[\frac{1}{2\sigma^2}\|x-F(\theta_c+\delta)\|_2^2+\frac{\lambda}{2}\|\delta\|_2^2\right].
\end{equation}

Reported nonlinear adaptation scores are the lowest objective values observed during the prescribed finite-budget Adam fitting procedure; Eq \ref{eq:energy} defines the ideal optimization objective.
The first term rewards reconstruction; the second penalizes movement from the prior. Here $\sigma^2>0$ scales reconstruction error and $\lambda>0$ sets the movement penalty. ``Cheap'' means a low value of this objective, not necessarily fewer optimizer steps. Selecting the class under which an observation has lowest energy gives an adaptation-based classifier, somewhat analogous to compression-based classifiers which check under which encoding a new observation has the smallest representation. However, our purpose is not to build a competitive classifier but rather to use this task to study the network's response to a new target.

Let $\mu_c=F(\theta_c)$, $r_c=x-\mu_c$, and $J_c=\partial F/\partial\theta|_{\theta_c}$. The Jacobian $J_c$ maps a small weight change to its predicted output change. Its \emph{tangent kernel}, $K_c=J_cJ_c^\top$, describes the output directions that these weight changes make accessible. Replacing the network by $F(\theta_c+\delta)\approx\mu_c+J_c\delta$ gives the exact minimum of the linearized objective:
\begin{equation}
\label{eq:tangent}
E_c^{\mathrm{lin}}(x)=\tfrac12 r_c^\top C_c^{-1}r_c,
\qquad C_c=\sigma^2 I+\lambda^{-1}K_c.
\end{equation}
This is a \emph{Mahalanobis distance}. Under this metric, different output directions are weighted differently. Directions with large kernel eigenvalues are cheaper to change along. Thus the local description is $(\mu_c,K_c)$, the initial reconstruction together with its accessible changes, rather than $\mu_c$ alone. Eq. \ref{eq:tangent} uses the actual finite network, not an infinite-width approximation. It is exact for the linearized optimization problem; agreement with finite-step nonlinear fitting must be measured. Appendix~\ref{app:full-derivation} gives the derivation, loss normalization, and dependence on the parameter-movement metric.

\subsection{Encoding a sequence by continuing the previous fit}
\label{sec:framework-paths}

We encode a sequence of physical fields by repeatedly fitting the same network, using the weights fitted to one frame as the starting point for the next. 

This construction is motivated by settings where a single observed frame does not determine the relevant dynamics. Sequential fitting gives the representation a simple way to carry information from earlier observations forward, without adding a separately trained recurrent model.

We compare these sequential weight trajectories with fitting every frame independently and with using the raw field sequence directly, then test how much information about the underlying dynamics each representation retains.
For observed frames $u_1,\ldots,u_T$, let $\operatorname{Adapt}(\theta,u)$ run a fixed fitting procedure initialized at $\theta$. Starting from a shared prior $\bar\theta$, we compare
\begin{equation}
\label{eq:seq}
\begin{aligned}
\text{Independent:}\quad &\theta_t^{\mathrm{ind}}=\operatorname{Adapt}(\bar\theta,u_t),\\
\text{Sequential:}\quad &\theta_t^{\mathrm{seq}}=\operatorname{Adapt}(\theta_{t-1}^{\mathrm{seq}},u_t),
\qquad \theta_0^{\mathrm{seq}}=\bar\theta.
\end{aligned}
\end{equation}
Independent fitting starts over for every frame. Sequential fitting carries the previous weights forward. Within a comparison, both use the same architecture and fitting budget. The optimizer state is reset at every frame, so information passes between fits through the weights and not through retained Adam momentum.

\paragraph{Reading out physical regimes.}
We apply the same classification procedure to raw-frame sequences, independent-weight sequences, and sequential-weight sequences. Features are standardized and compressed by principal component analysis (PCA), fitted only on support data. In the resulting coordinates $z_t$, each class learns a ridge-regularized linear next-step model,
\begin{equation}
\label{eq:readout}
z_{t+1}-z_t\approx A_c z_t+b_c,
\qquad
S_c=\frac{1}{T-1}\sum_{t=1}^{T-1}\|z_{t+1}-z_t-A_cz_t-b_c\|_2^2.
\end{equation}
The predicted class is the one which minimizes $S_c$. PCA rank and ridge strength are selected on validation data using the same protocol for each representation. All three methods receive the observation sequence. Separately, the wave-memory experiment uses only the final weights $\theta_T$ as the probe input.

\subsection{Local geometry sets memory timescales}
\label{sec:framework-retention}

The same local geometry that determines which output changes are easy to make also
determines how quickly earlier parameter changes are overwritten. 
Under gradient descent on a frozen Jacobian with half-squared error, a perturbation along tangent mode \(i\), with kernel eigenvalue \(\kappa_i\), retains the fraction 
\begin{equation}
\label{eq:retention}
\rho_i(s)=(1-\eta\kappa_i)^s
\end{equation}
after \(s\) steps. For \(0\leq\eta\kappa_i\leq1\), larger-\(\kappa_i\) modes decay faster.
Large-$\kappa_i$ modes are therefore easy to
change but quickly overwritten, while small-$\kappa_i$ modes change more slowly and
persist for longer. We test this prediction at sequentially fitted networks in
Section~\ref{sec:mechanism}.
This is a statement about the controlled linearized update, not the practical Adam encoder.

\paragraph{A closed-loop test of adaptation geometry}
\label{sec:holonomy-methods}

To test whether the current sampled output uniquely determines local adaptation geometry,
we make a network follow a closed loop of target fields. At each point we use the
minimum-norm parameter velocity that realizes the requested output velocity,
\begin{equation}
\label{eq:transport}
\dot\theta = J_\theta^+ \dot y .
\end{equation}
The final target is identical to the first. We therefore measure whether the network's
parameters and tangent kernel also return to their starting values. 
Full numerical checks, small-loop curvature diagnostics, and controls are given in
Appendix~\ref{app:holonomy-protocol}.

\FloatBarrier
\section{Experiments and results}
\label{sec:experiments}

We first test whether local geometry predicts adaptation and then whether sequential fitting retains useful history. We next connect these results through local retention dynamics. Finally, we include a controlled existence test showing that identical sampled outputs need not imply identical adaptation geometry. Appendices~\ref{app:static-protocol}--\ref{app:holonomy-protocol} contain the full protocols and numerical controls.

\subsection{Local geometry predicts nonlinear adaptation}
\label{sec:static}

\paragraph{Setup.}
We learn class priors for MNIST, Fashion-MNIST, and KMNIST \citep{lecun1998mnist,xiao2017fashionmnist,clanuwat2018kmnist}. The eight settings combine dataset, number of labeled support images, and architecture choices. We use a  sinusoidal network, SIREN \citep{sitzmann2020siren} and a structured M-layer  architecture \citep{fischbacher2020mlayer}. Justification for these architectures can be found in Appendix~\ref{app:inr-architectures}. Each uses three data seeds and held-out queries. Tuning is done on MNIST-30 (the low-data MNIST setting with 30 labeled support images in total, balanced as three examples per digit class) and frozen elsewhere. We compare a prior's own kernel with a kernel shared across classes and with kernels reassigned from other classes. Reassignment keeps each initial reconstruction $\mu_c$ fixed \emph{in the scoring formula}; it is not an edit of the network's actual weights.

\begin{table}[t]
\caption{\textbf{The correct local geometry predicts nonlinear adaptation.}
Correlation measures similarity in score rankings; agreement measures selection of the same prior.
Shared uses the mean kernel; Match uses each prior's own kernel; Shuffle reassigns kernels while fixing initial reconstructions (100 derangements).
$\Delta$Adapt. is nonlinear-minus-static classification accuracy in percentage points, a separate measure of utility.
Entries are pooled held-out metrics rather than seed means; paired uncertainty
for the method differences is reported in Appendix~\ref{app:static-replication}.}
\label{tab:static}
\centering
\footnotesize
\setlength{\tabcolsep}{3.0pt}
\renewcommand{\arraystretch}{1.08}
\begin{tabular}{lrrrrrrr}
\toprule
& & \multicolumn{3}{c}{Spearman $\rho$} & \multicolumn{3}{c}{Decision agreement (\%)} \\
\cmidrule(lr){3-5}\cmidrule(lr){6-8}
Setting & $\Delta$Adapt. & Shared & Match & Shuffle & Shared & Match & Shuffle \\
\midrule
MNIST-30 M-layer    & +12.2 & 0.829 & \textbf{0.939} & 0.690 & 73.3 & \textbf{84.4} & 61.9 \\
MNIST-100 M-layer   &  +5.3 & 0.886 & \textbf{0.955} & 0.770 & 75.7 & \textbf{88.3} & 58.9 \\
Fashion-100 M-layer &  +2.0 & \textbf{0.918} & 0.906 & 0.791 & 77.0 & \textbf{89.3} & 61.9 \\
KMNIST-100 M-layer  &  +0.0 & 0.884 & \textbf{0.935} & 0.771 & 65.0 & \textbf{79.7} & 47.7 \\
\midrule
MNIST-30 SIREN      &  +4.4 & 0.618 & \textbf{0.873} & 0.327 & 38.9 & \textbf{57.8} & 11.0 \\
MNIST-100 SIREN     & $-$15.3 & 0.733 & \textbf{0.853} & 0.466 & 33.0 & \textbf{58.3} &  6.6 \\
Fashion-100 SIREN   & $-$19.0 & 0.633 & \textbf{0.825} & 0.429 & 29.0 & \textbf{66.7} &  5.3 \\
KMNIST-100 SIREN    & $-$22.3 & 0.686 & \textbf{0.790} & 0.533 & 12.3 & \textbf{48.3} &  6.0 \\
\bottomrule
\end{tabular}
\end{table}

\paragraph{Results.}
Table~\ref{tab:static} separates two prediction measures: Spearman correlation asks whether candidate scores have similar rankings to those produced by full adaptation; decision agreement asks whether they choose the same prior. On MNIST-100, the M-layer's own kernel gives $0.955$ correlation and $88.3\%$ decision agreement, compared with $0.886$ and $75.7\%$ for shared geometry. Across 27 independently learned prior sets per architecture, the own-kernel advantage over shared geometry persists on MNIST-100 (Appendix~\ref{app:static-replication}).

The own-kernel improvement is also stable under resampling. On MNIST-100 with
the M-layer, the own-versus-shared gain is $+0.069$ in Spearman correlation
(95\% CI $[0.061,0.078]$) and $+12.7$ percentage points in decision agreement
(95\% CI $[8.7,16.7]$). Full query, fold, and independent-prior uncertainty
analyses are reported in Appendix~\ref{app:static-replication}.

The most consistent result is the reassignment control. Giving a prior another class's kernel reduces both measures in all settings. On MNIST-100, correlation falls from $0.955$ to $0.770$ for the M-layer and from $0.853$ to $0.466$ for SIREN. Equalizing kernel traces (their total scale) preserves the gap in every setting. Thus the pairing between a prior's reconstruction and its local geometry matters, beyond just the total kernel scale.

Kernels also transfer across independent fits: same-class donors predict nonlinear adaptation rankings better than different-class donors for both architectures, including after trace normalization, while own kernels remain strongest (Appendix~\ref{app:independent-kernel-transfer}).

Thus, \emph{local geometry helps predict which adaptations a fitted network favors.}
The connection to the sequence experiments is not that the image-class kernels transfer to physical systems; it is that the same local object, the tangent spectrum of the currently fitted network, appears in a local model of its response to the next fitting update.
We now ask what repeated applications of those updates retain over time.

\subsection{Different histories remain distinguishable at the same observation}
\label{sec:collision}
\label{sec:paths}

To test memory directly, we construct pairs of wave trajectories with different histories but exactly the same final observed displacement. Their final velocities are opposite. A deterministic classifier given only that final observation must make the same prediction for both members of a pair. With both opposite labels present, it is correct on exactly one: accuracy is exactly $50\%$. Independent fitting obeys the same bound when identical observations use the same initialization and fitting.

\paragraph{Setup.}
The experiment uses $3\times3$ data and encoder seeds and 384 unique physical pairs. The 2,241-parameter SIREN encoder is selected using reconstruction alone and frozen before hidden-velocity labels are accessed. We verify exact observation equality and bit-identical independent encodings (Appendix~\ref{app:collision}).

\begin{figure}[t]
    \centering
    \includegraphics[width=\linewidth]{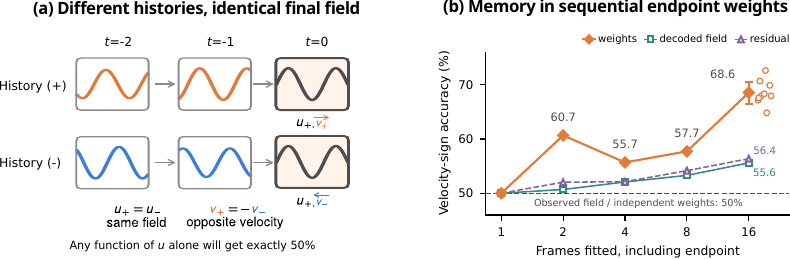}
    \caption{\textbf{Different histories stay distinguishable at the same observed endpoint.}
    \textbf{(a)} Paired wave histories differ in their earlier frames but end at exactly the same displacement field with opposite hidden velocities.
    \textbf{(b)} Velocity-sign accuracy of a linear probe on sequential endpoint weights as a function of how many frames were fitted before the endpoint; the observed field and independently fitted weights sit at chance by construction, and the field decoded from the sequential weights and its residual reach 55.6\% and 56.4\%. Dots at 16 frames are the nine seed cells; the whisker is the paired-bootstrap 95\% interval.}
    \label{fig:collision-summary}
\end{figure}

\paragraph{Results.}
Linear probes on sequential endpoint weights reach $68.6\%$ mean sign accuracy, with a minimum of $64.8\%$ across cells. The gain over independent weights is positive in all nine settings, with paired bootstrap 95\% interval $[16.5,20.6]$ percentage points. Velocity regression gives $R^2=0.22$, versus zero for the current-observation controls. The field decoded from the sequential endpoint reaches 55.6\%, and the endpoint reconstruction residual reaches 56.4\%. Thus the sequential weights make historical information more accessible to this probe, but the experiment does not establish that this information is entirely absent from the reconstruction (Figure~\ref{fig:collision-summary}).
The endpoint recovery is strongest with the full 16-frame history, although it does not increase monotonically with history length, and it is essentially unchanged across per-frame budgets of 25 to 500 steps (Appendix~\ref{app:wave-followup}). 

\subsection{Sequential weights improve physical-regime classification}
\label{sec:seq-results}

We encode a physical trajectory by repeatedly fitting the same small neural field, using the weights fitted to one frame as the starting point for the next. We then ask whether the resulting weight sequence makes the underlying dynamics easier to identify than either the raw field sequence or a sequence of networks fitted independently to each frame.

If history contains information about unresolved state or unknown dynamics, then a representation with memory may have something useful to retain.
Our first question is therefore whether sequential fitting turns the fitting path into a useful state variable: does carrying the previous network forward make the observed dynamics easier to distinguish than representing each frame independently?
The source of historical information differs across these tasks: the wave experiment omits velocity, and the shear-flow and \GS{} fields are spatially reduced, so both are partial observations in the \MZ{} sense, whereas \AC{} observes $u$ on the simulated grid and treats the reaction rate $a$ as unknown, which makes it a parameter-identification task rather than a reduced-state one.

\paragraph{Setup.}
We use three physical systems with different sources of temporal information. \AC{} is a scalar reaction--diffusion system whose four classes are different reaction rates \citep{allen1979cahn}. \GS{} is a two-species reaction--diffusion system with six pattern regimes \citep{pearson1993grayscott,ohana2024well}. Shear flow is a fluid simulation from The Well with four Reynolds-number classes \citep{ohana2024well}. We apply the readout of Equation~\ref{eq:readout} to each system. The \GS{} and shear-flow fields are spatially reduced and therefore partially observed in the \MZ{} sense, whereas \AC{} observes $u$ on the simulated grid and treats the reaction rate $a$ as unknown, making it a parameter-identification task rather than a reduced-state one. Encoders have 337 to 388 parameters. \AC{} and \GS{} cross three data seeds with three encoder seeds.

\begin{table}[t]
\caption{\textbf{Sequential fitting makes weight sequences more useful for classifying dynamics.}
Accuracy (\%) as seed mean / score ensemble; an ensemble standardizes each query's class-score vector within a run and averages across runs before deciding. Raw + delays selects among stacks of up to 16 raw frames in total: the current frame and at most 15 preceding frames. It reports means over three data seeds for \AC{} and \GS{}, and one deterministic run for shear flow, where validation selects $H=1$ (no added history). The delay baseline uses the primary evaluation, whose \GS{} raw comparator is 56.3 rather than the diagnostic 57.0 shown here. The final column is the sequential seed mean minus the best other mean or deterministic result; differences are computed before rounding. Chance is 25\% for \AC{} and shear flow and 16.7\% for \GS{}. \AC{} and \GS{} cross three data seeds with three encoder seeds; shear flow uses one support split and three encoder seeds. }
\label{tab:seq}
\centering\small
\setlength{\tabcolsep}{4pt}
\renewcommand{\arraystretch}{1.1}
\begin{tabular}{lccccc}
\toprule
System & Raw fields & \shortstack{Independent\\weights} &
\shortstack{Sequential\\weights} & \shortstack{Raw +\\delays} & \shortstack{Seq. minus\\best}\\
\midrule
\AC{}     & 42.1 / 48.8 & 31.9 / 37.5 & \textbf{64.4 / 66.2} & 65.4 & $-1.0$ \\
\GS{}     & 57.0 / 53.3 & 59.0 / 55.6 & \textbf{64.9 / 65.6} & 59.3 & $+5.7$ \\
Shear flow (45 support/class) & 27.7        & 28.6 / 33.0 & \textbf{34.8 / 44.6} & 27.7 ($H=1$) & $+6.2$ \\
\bottomrule
\end{tabular}
\end{table}

\paragraph{Results.}
Sequential weights outperform both raw fields and independent weights on all three systems in Table~\ref{tab:seq}. The stronger raw-history baseline separates the cases. On \GS{}, sequential weights reach $64.9\%$ seed-mean accuracy versus $59.3\%$ for validation-selected raw delays; on shear flow they reach $34.8\%$ versus $27.7\%$, and validation selects no added raw history ($H=1$). On \AC{}, by contrast, explicit raw delays match the sequential result ($65.4\%$ versus $64.4\%$). Thus the clearest sequential-weight advantage is on the reduced-field \GS{} and shear-flow tasks, while \AC{} provides a useful boundary case. Full seed-level and per-class results are in Appendix~\ref{app:seq-replication}.

\paragraph{What the controls establish.}
On shear flow, applying the same readout to fields decoded from sequential weights gives $26.8\%$, compared with $44.6\%$ for the weights themselves. Removing the extra penalty for moving from the preceding fit does not remove the gain ($45.5\%$ with $\beta=0$). Additional intermediate checkpoints and an objective encouraging different fitting routes to produce the same representation also fail to improve performance.
The validation-selected raw-delay baseline closes the sequential advantage on \AC{}; the advantage remains on \GS{} and shear flow, where the latter search selects no added history (Table~\ref{tab:seq}, Appendix~\ref{app:seq-controls}).

\subsection{The same geometry controls adaptation and forgetting}
\label{sec:mechanism}

\paragraph{The correct Jacobian matters in a diagnostic update model, on one system.}
\label{sec:tangent-intervention}
We approximate each transition by a regularized tangent update toward the next frame, then substitute a Jacobian from the wrong transition. On \GS{}, matched updates yield $46.5\%$ mean classification accuracy versus $30.1\%$ for mismatched updates, positive in all nine cells. The same intervention fails on shear flow, where matched and mismatched updates are both near chance (ensemble accuracies $26.8\%$ versus $28.6\%$), so the diagnostic supports a system-dependent role for Jacobian identity rather than a general mechanism (Appendix~\ref{app:tangent-protocol}).

\paragraph{Reachability and forgetting share one spectrum.}
\label{sec:retention}
The tangent kernel first appeared in Section~\ref{sec:static} as a measure of
\emph{reachability}: directions with larger kernel eigenvalues can be produced by smaller
parameter changes. The same eigenvalues also determine how quickly information in those
directions is overwritten by subsequent fitting.
To see this, consider gradient descent on a frozen local linearization. A perturbation along
a mode with curvature $\kappa_i$ retains the fraction \( \rho_i(s)=(1-\eta\kappa_i)^s \)
after $s$ steps of learning rate $\eta$. Thus the spectrum induces a set of local memory
timescales. Large-$\kappa_i$ modes are easy to write but are overwritten quickly; small-$\kappa_i$ modes are harder to change but persist for longer. In this local model, \emph{adaptability and memory are therefore controlled by the same geometry}.

This prediction is quantitatively accurate under the controlled updates we test.
At saved sequential endpoints, predicted and measured mode-wise retention have median
rank correlations of $0.986$ on \AC{} and $0.945$ on \GS{}, with
log-retention slopes near the predicted value of one. After 16 steps, the slowest quartile of modes retains $0.998/0.999$ of its perturbation, while the fastest quartile retains only $0.259/0.579$ (Appendix~\ref{app:retention}).
This establishes the predicted local retention timescales, but not that the slow instantaneous modes carry the useful classification memory: output-matched selective erasure does not consistently damage performance more when slow modes are removed (Appendix~\ref{app:erasure}).

\begin{table}[t]
\caption{\textbf{More fitting improves reconstruction but eventually hurts the
history-bearing representation.} Validation results for sequential fitting. Reconstruction
error decreases monotonically with the number of fitting steps, while regime-classification
accuracy peaks at an intermediate budget.}
\label{tab:budget-main}
\centering
\small
\setlength{\tabcolsep}{5pt}
\begin{tabular}{lrrr}
\toprule
System & Steps/frame & Seq.\ MSE & Seq.\ accuracy \\
\midrule
\AC{} & 50   & 0.059 & 64.6 \\
             & 200  & 0.043 & \textbf{69.4} \\
             & 1000 & \textbf{0.035} & 60.1 \\
\midrule
\GS{} & 50   & 0.679 & \textbf{69.2} \\
             & 100  & 0.599 & 68.8 \\
             & 1000 & \textbf{0.395} & 64.1 \\
\bottomrule
\end{tabular}
\end{table}

The practical encoder shows the corresponding budget trade-off. Increasing the fitting
budget monotonically improves reconstruction, but classification peaks at an intermediate
number of updates and then declines (Table~\ref{tab:budget-main}). This is what we would
expect if finite-budget fitting acts as a mode-dependent temporal filter: each new frame is
written strongly into fast tangent modes while information in those modes is rapidly
overwritten, whereas slower modes preserve effects of earlier frames for longer.
The wave experiment does not show this washout over its tested range of 25 to 500 steps per frame (Appendix~\ref{app:wave-followup}); its probe is supervised, its encoder is about six times larger, and the velocity signal may occupy slow modes, so the budget trade-off is established for the regime tasks and open for the wave task.

Equation~\ref{eq:retention} is exact for the controlled linearized update rather than the practical Adam encoder. Differentiating through the actual Adam fits passes one-frame numerical checks but becomes unstable over an eight-frame horizon, so we do not perform the corresponding causal erasure test; which directions carry useful finite-horizon memory therefore remains open (Appendix~\ref{app:fh}).

\label{sec:holonomy}
\paragraph{Controlled loop at fixed sampled outputs.} In a controlled transport experiment on a complete \(12\times12\) grid, the median relative output error after a closed loop is \(3\times10^{-14}\) across nine cells at the two smallest loop scales, while the parameters and tangent kernel remain changed. This is an existence result at fixed sampled outputs; ordinary SGD and Adam do not reproduce the same transport signatures (Appendix~\ref{app:holonomy-protocol}).

\section{Discussion}
\label{sec:discussion}

What should we evaluate in a fitted representation? Our main results point to two properties beyond reconstruction: how a fitted network responds to further updates, and what repeated fitting retains from earlier observations. The tangent spectrum connects these two locally by determining both reachability and forgetting timescales. The controlled-loop experiment adds a separate geometric observation: even fixing the sampled reconstruction does not, in general, fix this local adaptation geometry.

\section{Limitations}
\label{sec:limitations}

The main unresolved question is how completely the local tangent picture explains the
memory produced by practical sequential fitting. The retention rule is exact for a frozen
linearization under controlled gradient descent, while the practical encoder uses nonlinear
Adam updates. Our results connect tangent geometry to memory timescales, but do not
establish that the instantaneous tangent modes are the causal carrier of the full
classification advantage. Consistent with this distinction, selective erasure does not identify the slow instantaneous modes as the useful-memory carrier, and a finite-horizon analysis through the actual Adam fits fails numerical prerequisites before its causal erasure test (Appendices~\ref{app:erasure} and~\ref{app:fh}).

The sequence experiments also measure usefulness under one deliberately matched
class-conditioned dynamics readout. They show that sequential weights expose useful
history to this readout, not that they outperform every possible classifier on the raw
sequence. Likewise, in the endpoint-collision experiment the final \emph{observations} are
identical, but the sequentially decoded fields are not perfectly identical; the result therefore
establishes retained history rather than completely output-independent hidden state.

Finally, the tangent geometry is defined jointly by the network, loss, and parameter-movement
metric, so its numerical form is parameterization-dependent. Broader architectures and
modalities will be needed to determine how generally the observed geometry--memory
relationship holds.

\section{Conclusion}
\label{sec:conclusion}

{Reconstruction is only one aspect of a neural-field representation.} A fitted network also has a measurable local response to change, and repeated fitting can leave useful history in its weights. The practical implication is to evaluate these properties rather than assume that better reconstruction necessarily gives a better representation. 

\subsection*{AI use statement}
In this work, we used generative AI tools to assist with programming, experimental design and execution, analysis of results, literature search, and drafting and editing the manuscript. We did not use generative AI tools to generate synthetic datasets or to prove mathematical claims. All AI-assisted code was reviewed and tested by the authors, AI-suggested references were checked against the original publications, and all derivations in the appendices were verified by hand. We take responsibility for the final content of this work, including text, claims, and artifacts produced with the aid of generative AI.

\subsection*{Ethics statement}
This work uses public image benchmarks, public physics simulations from The Well, and synthetic physical fields. It involves no human-subject experiments or personal data. We report unsuccessful as well as successful experiments, preserve held-out evaluation, and distinguish negative results from tests that could not meet numerical prerequisites.

\subsection*{Reproducibility statement}
The appendices document the data splits, architectures, fitting procedures, hyperparameter selection, readouts, uncertainty analyses, and numerical controls used for the reported experiments. They also distinguish held-out and validation evaluations and record failed or numerically ineligible tests.

\bibliographystyle{iclr2027_conference}
\bibliography{references}

\clearpage
\appendix

\section{Local adaptation: derivations and controls}
\label{app:full-derivation}

\subsection{The objective and its implemented normalization}
\label{app:notation}
The adaptation objective combines reconstruction error with the cost of moving away from a learned initialization. The more general form allows different parameter directions to have different movement costs:
\begin{equation}
\label{eq:app-general-energy}
\mathcal E_c(x)=\min_\delta\left[
\frac{1}{2\sigma^2}\|x-F(\theta_c+\delta)\|_2^2+
\frac12\delta^\top\Lambda_c\delta\right],\qquad\Lambda_c\succ0.
\end{equation}
The positive-definite matrix $\Lambda_c$ specifies the movement metric. The image experiments use the isotropic special case
\begin{equation}
\label{eq:app-implemented-energy}
E_c(x)=\min_\phi\left[\operatorname{MSE}(x,F(\phi))+
\beta\|\phi-\theta_c\|_2^2\right],\qquad\beta=10^{-3}.
\end{equation}
When the mean squared error averages over $m$ scalar targets, the two expressions agree for $\sigma^2=m/2$ and $\Lambda_c=\lambda I$ with $\lambda=2\beta$. For the five-field M-layer, $m=3920$, $\sigma^2=1960$, and $\lambda=0.002$. For the intensity-only SIREN, $m=784$, with the same conversion. The implemented nonlinear score is the minimum objective value encountered during finite-step fitting, not a guaranteed global minimum.

\subsection{Solving the linearized adaptation problem}
\label{app:tangent-proof}
Write $\mu_c=F(\theta_c)$, $r_c=x-\mu_c$, and $J_c=\partial F/\partial\theta|_{\theta_c}$. Substituting the local approximation $F(\theta_c+\delta)\approx\mu_c+J_c\delta$ gives
\[
Q_c(\delta)=\frac{1}{2\sigma^2}\|r_c-J_c\delta\|_2^2+
\frac12\delta^\top\Lambda_c\delta.
\]
For this derivation only, let $H_c=\Lambda_c+\sigma^{-2}J_c^\top J_c$ and $g_c=\sigma^{-2}J_c^\top r_c$. Setting the derivative to zero gives $\delta_c^*=H_c^{-1}g_c$ and
\[
\min_\delta Q_c(\delta)=\frac{1}{2\sigma^2}r_c^\top r_c-
\frac12g_c^\top H_c^{-1}g_c.
\]
Woodbury's matrix identity rewrites this expression in output coordinates:
\begin{equation}
\label{eq:app-mahalanobis}
E_c^{\mathrm{lin}}(x)=\tfrac12r_c^\top C_c^{-1}r_c,
\qquad C_c=\sigma^2 I+J_c\Lambda_c^{-1}J_c^\top.
\end{equation}
This calculation uses the Jacobian of the trained finite network. It does not assume that the network is infinitely wide or remains linear throughout nonlinear adaptation.

For $\Lambda_c=\lambda I$, diagonalize $K_c=J_cJ_c^\top$ as $U_c\operatorname{diag}(\kappa_{c,i})U_c^\top$. If $u_{c,i}$ is the corresponding output direction, then
\begin{equation}
\label{eq:app-spectral}
E_c^{\mathrm{lin}}(x)=\frac12\sum_i
\frac{(u_{c,i}^\top r_c)^2}{\sigma^2+\kappa_{c,i}/\lambda}.
\end{equation}
This makes the direction-dependent cost explicit. A residual aligned with a large kernel eigenvalue contributes less energy than an equally large residual along a weak direction. The prior output is the center of the local model; its kernel determines the shape of the surrounding equal-cost ellipsoids.

\subsection{Relation to the nonlinear objective near the prior output}
\label{app:secondorder}
The tangent energy is also the quadratic Taylor term of the nonlinear adaptation energy at the prior output $x=\mu_c$. Assume $F$ is twice continuously differentiable near $\theta_c$. At $r=x-\mu_c=0$, the minimizing displacement is zero, and the Hessian with respect to displacement is $H_c\succ0$. The implicit-function theorem gives a locally smooth minimizing displacement as a function of $r$.

Differentiating the objective after minimizing over displacement yields
\[
\nabla_r\mathcal E_c(\mu_c)=0,\qquad
\nabla_r^2\mathcal E_c(\mu_c)=
\sigma^{-2}I-\sigma^{-4}J_cH_c^{-1}J_c^\top=C_c^{-1}.
\]
Hence
\[
\mathcal E_c(\mu_c+r)=\tfrac12r^\top C_c^{-1}r+o(\|r\|_2^2).
\]
This is a local statement about the objective. It does not guarantee accurate predictions for large adaptations or for the final output of a finite-step optimizer.

\subsection{Shared, reassigned, and scale-normalized kernels}
\label{app:geometry-controls}
The static score uses only the residual from the prior output, $\|x-\mu_c\|_2^2$. The shared-kernel score uses $\bar K$, the mean kernel across classes, in every candidate's distance. This allows direction-dependent adaptation, but removes candidate-specific kernel identity.

For reassignment, choose a permutation $\pi$ in which no class keeps its own kernel. Such a permutation is a \emph{derangement}. We replace $C_c$ by
\[
C_{c,\pi}=\sigma^2I+\lambda^{-1}K_{\pi(c)},
\]
while keeping the prior output $\mu_c$ fixed. Each reported shuffled value averages 100 derangements. This is an intervention on the scoring rule, not a construction of a network with the reassigned kernel.

To test whether total kernel scale explains the effect, repeat the comparison after setting
\[
\widetilde K_c=K_c\frac{\bar t}{\operatorname{tr}(K_c)},
\qquad \bar t=\text{mean}_c\operatorname{tr}(K_c).
\]
Every kernel then has the same trace, while retaining its eigenvalue ratios and eigendirections. Persistence of the gap shows that total scale alone does not explain the matched-kernel advantage.

\subsection{Dependence on parameter coordinates}
\label{app:reparam}
The size of a weight change depends on the parameter coordinates used to describe it. For a locally invertible change of coordinates $\delta'=B\delta$, the Jacobian transforms as $J'=JB^{-1}$. To preserve the same movement cost, the metric must transform as $\Lambda'=B^{-\top}\Lambda B^{-1}$. The combination
\[
J'\Lambda'^{-1}J'^\top=J\Lambda^{-1}J^\top
\]
is then unchanged. The geometry is therefore defined by the network, loss, and movement metric together, rather than by the network function alone \citep{amari1998natural}. Fixing a Euclidean penalty in different coordinates need not describe the same adaptation problem.

\section{Image experiments: protocols and extended results}
\label{app:static-protocol}

\subsection{Data splits and hyperparameter selection}
\label{app:static-data}
MNIST support sets contain 20, 30, or 100 balanced labels, or 2--10 examples per class. Data-selection seeds are 0--2. The 20-, 30-, and 100-label studies use respectively 2, 3, and 10 leave-one-per-class folds per seed. A query is excluded from the support set used to fit its own class prior.

A fixed 5,000-image MNIST test split is used only for the frozen 20- and 30-label evaluations, after the score and movement penalty $\beta$ have been fixed. It is not loaded for score selection, tangent analysis, kernel reassignment, or bootstrap analysis. Fashion-MNIST and KMNIST use 100 balanced labels, three seeds, and 10 folds per seed, giving 300 cross-validation queries per dataset.

The movement penalty $\beta=0.001$ was selected in a separate 20-label MNIST study. The M-layer tangent scorer and protocol selected on MNIST-30 are reused on MNIST-100 and both transfer datasets. For SIREN, a 36-configuration grid on MNIST-30 varies width over $\{32,64\}$, frequency $\omega$ over $\{10,30\}$, and Reptile and query learning rates over $\{5\times10^{-4},10^{-3},5\times10^{-3}\}$. It selects width 32, $\omega=10$, and both learning rates $5\times10^{-4}$. This configuration is frozen for the other three SIREN settings.

\begin{table}[htbp]
\caption{\textbf{Architecture and fitting settings for the image experiments.}}
\label{tab:app-static-protocol}
\centering
\small
\setlength{\tabcolsep}{5pt}
\renewcommand{\arraystretch}{1.08}
\begin{tabular}{lll}
\toprule
Component & Five-field M-layer & Intensity-only SIREN \\
\midrule
Outputs & $5\times28\times28=3920$ & $28\times28=784$ \\
Adapted parameters & 913 & 2{,}241 \\
Prior initialization & Common within each fold & Common within each fold \\
Reptile inner steps & 8 Adam steps & 8 steps \\
Reptile inner learning rate & 0.04 & $5\times10^{-4}$ \\
Reptile outer steps & 1{,}000 & 1{,}000 \\
Outer interpolation & 0.1, constant & 0.1, constant \\
Query learning rate & 0.04 & $5\times10^{-4}$ \\
Regularization & $\beta=0.001$ & $\beta=0.001$ \\
\bottomrule
\end{tabular}
\end{table}

The M-layer uses three structured branches to decode five target fields derived from intensity, Sobel, and distance-transform channels. Its Jacobians are evaluated in exact block form. The SIREN is a conventional two-dimensional sinusoidal coordinate network with scalar output. All ten priors within a fold start from the same base parameter vector.

At inference, each query is fitted independently from every frozen class prior. Every candidate receives fresh optimizer state, the same penalty, and at most 2,000 steps with early-stopping patience 200. The score is the lowest observed value of Equation~\ref{eq:app-implemented-energy}; the adapted copy is discarded. No separate network is trained to classify the adapted image weights.

\subsection{Choice of INR architectures}
\label{app:inr-architectures}

We use SIRENs and M-layers to test whether the adaptation-geometry effect persists across two substantially different parameterizations that can represent periodic structure. SIREN uses sinusoidal activations, while the M-layer’s matrix-exponential nonlinearity can also represent oscillatory structure
\citep{sitzmann2020siren}
\citet{fischbacher2020mlayer}.

\subsection{Prediction metrics and uncertainty}
\label{app:static-stats}
Spearman correlation is computed over complete query--class score vectors. Decision agreement is the fraction of queries for which a comparison score selects the same prior as nonlinear adaptation. Both measure prediction of the adaptation procedure; neither is ground-truth classification accuracy.

Paired query bootstraps use 20,000 replicates, resampling queries within data-seed and true-class strata while retaining complete class-score vectors. Fold bootstraps use 10,000 replicates for the original M-layer MNIST-30 and MNIST-100 comparisons, and 20,000 for the SIREN comparisons and the M-layer Fashion-MNIST and KMNIST comparisons. The independent-prior experiment repeats meta-training across 27 prior sets per architecture: three data seeds, three folds, and three Reptile seeds. Query and fold intervals condition on fitted priors, while the latter experiment also varies prior fitting.

\subsection{Replication and the distinction from classification accuracy}
\label{app:static-replication}
\begin{table}[htbp]
\caption{\textbf{Replication of the own-kernel advantage over shared geometry.} $\Delta\rho$ is the correlation difference; $\Delta$Agree is the decision-agreement difference in percentage points. Intervals are 95\% confidence intervals. The last two rows refit the priors.}
\label{tab:app-bootstrap}
\centering
\scriptsize
\setlength{\tabcolsep}{3.2pt}
\renewcommand{\arraystretch}{1.08}
\begin{tabular}{lllrrrr}
\toprule
Setting & Replication unit & Comparisons & $\Delta\rho$ & 95\% CI & $\Delta$Agree & 95\% CI \\
\midrule
MNIST-30 M-layer & Query bootstrap & 900 pairs & +0.110 & [0.093, 0.130] & +11.1 & [3.3, 18.9] \\
                 & Fold bootstrap  & 900 pairs & +0.110 & [0.090, 0.132] & +11.1 & [2.2, 18.9] \\
MNIST-100 M-layer & Query bootstrap & 3{,}000 pairs & +0.069 & [0.061, 0.078] & +12.7 & [8.7, 16.7] \\
                  & Fold bootstrap  & 3{,}000 pairs & +0.069 & [0.060, 0.078] & +12.7 & [8.3, 17.0] \\
MNIST-100 M-layer & 27 prior sets & 2{,}700 pairs & +0.069 & [0.048, 0.095] & +13.7 & [5.6, 23.7] \\
MNIST-100 SIREN   & 27 prior sets & 2{,}700 pairs & +0.134 & [0.096, 0.192] & +33.0 & [22.6, 43.3] \\
\bottomrule
\end{tabular}
\end{table}

\begin{table}[htbp]
\caption{\textbf{The reassignment effect persists after equalizing total kernel scale.} Each kernel is rescaled to a common trace. Shuffle averages 100 derangements; decision agreement is in percent.}
\label{tab:app-trace}
\centering
\small
\setlength{\tabcolsep}{4.5pt}
\begin{tabular}{lrrrr}
\toprule
& \multicolumn{2}{c}{Spearman $\rho$} & \multicolumn{2}{c}{Decision agreement} \\
\cmidrule(lr){2-3}\cmidrule(lr){4-5}
Setting & Match & Shuffle & Match & Shuffle \\
\midrule
MNIST-30 M-layer    & 0.943 & 0.698 & 82.2 & 62.6 \\
MNIST-100 M-layer   & 0.957 & 0.770 & 89.0 & 58.5 \\
Fashion-100 M-layer & 0.909 & 0.797 & 87.3 & 62.3 \\
KMNIST-100 M-layer  & 0.935 & 0.778 & 79.0 & 48.5 \\
\midrule
MNIST-30 SIREN      & 0.841 & 0.362 & 64.4 & 10.9 \\
MNIST-100 SIREN     & 0.840 & 0.503 & 60.7 &  6.4 \\
Fashion-100 SIREN   & 0.817 & 0.443 & 61.3 &  6.0 \\
KMNIST-100 SIREN    & 0.772 & 0.554 & 48.0 &  6.4 \\
\bottomrule
\end{tabular}
\end{table}

For SIREN, using each prior's own kernel rather than the shared kernel improves both prediction metrics in all four settings. Correlation gains are $0.255$, $0.120$, $0.192$, and $0.104$ on MNIST-30, MNIST-100, Fashion-100, and KMNIST-100; decision-agreement gains are $18.9$, $25.3$, $37.7$, and $36.0$ percentage points. Paired query and fold intervals exclude zero for every setting. The final three are frozen evaluations.

Fashion-MNIST with the M-layer limits the shared-kernel comparison: the own kernel has slightly lower pooled correlation ($0.906$ versus $0.918$; paired 95\% interval for the difference $[-0.0227,-0.0019]$). Its decision agreement is nevertheless higher ($89.3\%$ versus $77.0\%$), and its own kernel outperforms reassigned kernels. The result shared by all eight settings is the benefit of the correct mean--kernel pairing over reassignment, not an improvement over shared geometry on every measure.

The reassignment effect also survives equalized traces (Table~\ref{tab:app-trace}). On MNIST-30 with the M-layer, none of the 100 unnormalized derangements matches the correct pairing in accuracy, correlation, or decision agreement. The plus-one Monte Carlo randomization value is $p=0.0099$ for each measure.

The replicated result is prediction of nonlinear adaptation, not a superior image classifier. In the low-label comparisons, direct-image CNNs remain stronger: $82.8\%$ and $89.9\%$ at 20 and 30 labels, versus $72.3\%$ and $66.4\%$ for nonlinear adaptation.

\subsection{Independent-refit kernel transfer}
\label{app:independent-kernel-transfer}

Both architectures show
\[
\textbf{Own} > \textbf{Same-class donor} > \textbf{Shared} > \textbf{Different-class donor}
\]
for within-query rank correlation and decision agreement. The pattern persists after common-trace normalization. Thus, class-associated geometry transfers across independent fits, while some useful geometry remains specific to the recipient fit. Within-query rank correlation ranks the ten candidate priors for each query separately, whereas Table~\ref{tab:static} pools all query--class pairs; the shared kernel therefore scores lower here than in the main table.

The original-scale within-query Spearman results are:

\begin{table}[htbp]
\centering
\begin{tabular}{lrrrr}
\hline
\textbf{Architecture} &
\textbf{Own} &
\textbf{Same-class donor} &
\textbf{Different-class donor} &
\textbf{Shared} \\
\hline
M-layer & 0.9117 & 0.8539 & 0.5549 & 0.7385 \\
SIREN   & 0.7847 & 0.7142 & $-0.0802$ & 0.0917 \\
\hline
\end{tabular}
\end{table}

Same-class minus different-class Spearman is \( +0.2990\; [0.2775,\,0.3214] \) for M-layer and \( +0.7944\; [0.7705,\,0.8183] \) for SIREN. After trace normalization, the corresponding differences are \( +0.3104\; [0.2888,\,0.3329] \) and \( +0.7499\; [0.7246,\,0.7751], \) respectively. Own-minus-same-class gaps also exclude zero. 

Query bootstrap intervals resample complete query vectors within data-seed/true-class strata, carrying all fits and mappings together. Intervals condition on the saved fits; recipient/donor variation is reported separately. The conclusions concern the tangent scoring surrogate, with the actual nonlinear learner held fixed.

\subsection{Boundary checks for the tangent score}
\label{app:static-negatives}
\paragraph{Predicting scores does not mean predicting final reconstructions.}
For nonlinear endpoint $\phi_c^*$, define
\[
\epsilon_{\mathrm{lin}}=
\frac{\|F(\phi_c^*)-F(\theta_c)-J_c(\phi_c^*-\theta_c)\|_2}
{\|F(\phi_c^*)-F(\theta_c)\|_2+\epsilon}.
\]
On MNIST-30, means range from $0.66$ to $0.88$ and medians from $0.59$ to $0.87$ across subsets of query--class pairs. The tangent model preserves useful relative score structure without accurately reproducing the final adapted output.

\paragraph{The energy is not calibrated probabilistic evidence.}
Adding the Gaussian-evidence term $\frac12\log\det C_c$ performs poorly under the implemented metric and noise model (10.0\% classification accuracy and $-0.187$ correlation with nonlinear energy). The positive result therefore concerns the quadratic score, not a calibrated evidence interpretation.

\section{Shear-flow split integrity}
\label{app:priors-pde}

Preliminary experiments exposed duplicate initial conditions in the locally packaged shear-flow data. The final support, validation, and test manifests used for the reported results were rebuilt to be disjoint in canonical initial-condition identity.

\section{Sequential encoding: protocols, controls, and replication}
\label{app:seq-protocol}

\subsection{Systems, splits, and encoders}
\label{app:seq-systems}
\paragraph{Shear flow.}
The task has four Reynolds classes. Support is matched across nuisance variables: each class has the same 45 combinations of Schmidt number and initial-condition identity. This gives 180 support trajectories and 675 transitions per class over 16 source frames. 
The four normalized channels are bilinearly downsampled to a $32\times64$ grid with \texttt{align\_corners=False}.
Validation and held-out test each contain 112 trajectories, disjoint in initial condition.

The encoder is a width-16, one-hidden-layer SIREN with $\omega=10$ and 388 parameters. Each frame receives 50 fresh-Adam steps, starting from one shared, label-free Reptile prior for independent encoding or the preceding endpoint for sequential encoding. The initial run uses $\beta=10^{-3}$. A fixed coordinate schedule avoids adding sampling noise to the apparent temporal motion.

\paragraph{\AC{}.}
The simulated field satisfies
\[
\partial_tu=0.08\,\Delta u+a(u-u^3)
\]
on a periodic $32\times32$ grid, with reaction-rate classes $a\in\{0.5,1,2,4\}$. Each data seed has 12 support and 8 validation initial conditions per class, plus 20 fresh test initial conditions per class. Three data seeds are crossed with three encoder seeds. The width-16 SIREN has 337 parameters, uses 50 Adam steps per frame at learning rate $3\times10^{-3}$, and starts from a shared label-free Reptile prior. There is no movement penalty ($\beta=0$).

\paragraph{\GS{}.}
The Well supplies six pattern regimes: bubbles, gliders, maze, spirals, spots, and worms. The two-channel fields are area-downsampled to $32\times32$. The reported experiment uses disjoint blocks within the training files: support initial conditions 0--79, with 64 used per class; validation 100--119; an unused guard block 120--144; and a held-out result block 145--159. It uses frames 500--515, a 354-parameter SIREN, 200 Reptile steps, and 100 fresh-Adam steps per frame at learning rate $3\times10^{-4}$ with $\beta=0$. 
For sequential weights, the selected dynamics rank is 64 with zero ridge penalty.
Hyperparameters are selected on \GS{} validation initial conditions, prioritizing the worst encoder seed, and the held-out block is opened once.

\subsection{Matched readout and reporting conventions}
\label{app:seq-readout}
Raw frames and flattened network weights are standardized and mapped to a whitened PCA representation fitted only on support data. Each class fits a ridge-regularized linear model of successive differences, as in Equation~\ref{eq:readout}. The mean squared next-step residual is the class score. PCA rank and ridge strength are selected on validation data rather than the result split.

A seed mean averages classification accuracies across fitted data/encoder cells. A score ensemble standardizes each query's class-score vector within each run, averages these vectors across runs, and then makes the class decision. Shear flow has one data split and three encoder runs; \AC{} and \GS{} cross three data seeds with three encoder seeds.
Crossed seed cells measure different sources of variation and are not treated as independent datasets.

\GS{} is reported under two evaluation protocols on the same held-out block. The primary protocol reports raw fields at 56.3/57.8 and sequential weights at 64.9/65.6 (seed mean / score ensemble). Table~\ref{tab:seq} evaluates raw fields, independent weights, and sequential weights together, giving 57.0/53.3, 59.0/55.6, and 64.9/65.6, respectively. Because the sequential result is unchanged while the raw comparator differs, we keep the two protocols separate rather than pool them.

\subsection{Controls on what makes the sequential weights useful}
\label{app:seq-controls}
\begin{table}[htbp]
\caption{\textbf{Shear flow results and representation controls.} Accuracy (\%) is shown for the score ensemble and individual encoder seeds. Raw states and the validation-selected delay baseline are the same deterministic representation ($H=1$). Stored parameter counts include the dynamics readout for one encoder seed, excluding PCA arrays.}
\label{tab:app-re45}
\centering
\small
\setlength{\tabcolsep}{4.5pt}
\renewcommand{\arraystretch}{1.08}
\begin{tabular}{lcccr}
\toprule
Representation & Test (ensemble) & Encoder seeds & Macro-F1 & Stored params \\
\midrule
Raw states / selected delay ($H=1$) & 27.7 & n/a & 0.233 & 4{,}224 \\
Independent INR weights           & 33.0 & 33.9\,/\,23.2\,/\,28.6 & 0.235 & 4{,}612 \\
Sequential endpoints ($\beta=10^{-3}$) & 44.6 & 33.0\,/\,38.4\,/\,33.0 & 0.439 & \textbf{676} \\
Sequential endpoints ($\beta=0$)  & \textbf{45.5} & 33.9\,/\,38.4\,/\,33.9 & 0.439 & 676 \\
Full adaptation path (6 checkpoints) & 39.3 & 33.0\,/\,37.5\,/\,31.2 & 0.384 & 2{,}116 \\
Decoded sequential fields         & 26.8 & 23.2\,/\,24.1\,/\,31.2 & 0.225 & n/a \\
\bottomrule
\end{tabular}
\end{table}

On shear flow, sequential fitting improves reconstruction MSE from about $0.758$ for independent fits to $0.182$. This initially leaves better reconstruction as an alternative explanation. Decoding the sequential weights and applying the same readout gives only $26.8\%$, compared with $44.6\%$ for the weights. The tested decoded-field representation therefore does not reproduce the advantage; this does not prove that all possible readouts of decoded fields would fail.

Removing the movement penalty slightly improves the sequential ensemble to $45.5\%$. Concatenating six intermediate checkpoints per transition instead of using only frame endpoints lowers accuracy to $39.3\%$. The original $44.6\%$ ensemble corresponds to 50 correct predictions out of 112.

\paragraph{Encouraging different fitting routes to agree.}
A label-free objective reduces the MSE between direct and sequential fitting routes from $0.01131$ to $0.00957$, but with a matched four-step SGD adapter classification reaches only $35.7\%$, below the $45.5\%$ result for ordinary warm starts. Reducing this measured path dependence therefore does not improve the tested downstream task.

\paragraph{Explicit raw-frame delays.} We test whether explicit access to observation history reproduces the sequential-weight advantage. For \(H\in\{1,2,4,8,16\}\), we concatenate \(H\) raw frames, including the current frame, before support-fitted standardized PCA and apply the same class-conditioned linear dynamics scorer. Concatenation starts with the current frame and proceeds backward in time; missing early history repeats the first frame. Thus \(H=16\) includes at most 15 preceding frames. History length, PCA rank, and ridge strength are selected using the original validation procedure; \(H=1\) reproduces the current-raw decisions. On \AC{}, the selected \(H=16\) delay representation reaches \(65.4\%\), compared with \(42.1\%\) for current raw fields and \(64.4\%\) for sequential weights. On \GS{}, the selected \(H=4\) representation reaches \(59.3\%\), compared with \(56.3\%\) raw and \(64.9\%\) sequential. Thus explicit raw history closes the \AC{} gap under this readout, but does not reproduce the \GS{} result.
On shear flow, we use the same normalized $32\times64\times4$ fields, support-only preprocessing, and class-conditioned dynamics scorer. We search $H\in\{1,2,4,8,16\}$, PCA ranks $\{2,4,8,16,32,64\}$, and ridge strengths $\{0,10^{-4},10^{-3},10^{-2},10^{-1},1\}$ on the 112 validation trajectories, selecting by validation performance with fixed tie-breaking. The selected configuration is $H=1$, rank 32, and zero ridge. After freezing it and the PCA, test accuracy is $27.7\%$ ($31/112$), identical to the original raw baseline; sequential weights score $34.8\%$ as a mean and $44.6\%$ as an ensemble. Only the selected configuration and the $H=1$ control are evaluated on test; longer histories are compared on validation only. The historical test set had already been used in the earlier shear-flow studies.

For \AC{} and \GS{}, raw and delay results average three data-seed runs; the sequential result additionally crosses three encoder seeds per data seed. Shear flow has one fixed data split and one deterministic raw/delay run, compared with three existing sequential encoders. The delay search therefore does not provide three independent shear flow replications.

\subsection{Replication, classwise behavior, and forecasting}
\label{app:seq-replication}

The sequential-minus-raw difference is positive in every tested seed cell under the reported system-specific protocols. On shear flow, all three sequential encoder seeds beat the deterministic raw baseline. On \AC{}, all nine crossed data/encoder cells are positive, with improvements ranging from $5.0$ to $38.7$ points. On \GS{}, all nine cells are positive, with a minimum improvement of $5.6$ points.

\begin{table}[htbp]
\caption{\textbf{\GS{} held-out stability.} Accuracy (\%) and variation across encoder/data seeds for the primary held-out evaluation.}
\label{tab:app-graystable}
\centering
\small
\setlength{\tabcolsep}{5pt}
\renewcommand{\arraystretch}{1.08}
\begin{tabular}{lccccc}
\toprule
Representation & Mean & Worst seed & Seed SD & Range & Ensemble \\
\midrule
Raw fields & 56.3 & 54.4 & 1.39 & 3.3 & 57.8 \\
Sequential weights & \textbf{64.9} & \textbf{62.2} & \textbf{1.18} & 3.3 & \textbf{65.6} \\
\bottomrule
\end{tabular}
\end{table}

\GS{} remains uneven across classes despite its seed stability. Per-class sequential accuracies are maze $100.0\%$, spots $100.0\%$, worms $97.0\%$, spirals $85.9\%$, gliders $6.7\%$, and bubbles $0.0\%$, with seed standard deviations at most $6.4$. The aggregate therefore demonstrates a reproducible gain for this task while leaving substantial classwise failure modes.

On \GS{}, held-out one-step $R^2$ is near zero or negative across ranks for all representations: raw peaks near $-0.0004$, while independent and sequential weights peak near $+0.0014$. Thus the \GS{} classification gain does not rely on an accurate low-dimensional one-step forecast.

\section{Wave histories with exactly identical final observations}
\label{app:collision}

\subsection{Construction, exact bound, and encoder audit}

Each pair consists of two periodic two-dimensional wave trajectories with 80 real Fourier modes, common terminal displacement coefficients and wave speed, and opposite terminal velocity coefficients. Sixteen frames span $t\in[-0.75,0]$. At $t=0$, observed displacement fields are exactly equal and velocities exactly opposite.

The exact $50\%$ bound follows from the pairing. Any deterministic feature map of the terminal observation gives identical features to both members. A deterministic sign classifier must then assign the same label to both, making exactly one correct prediction per pair. The same argument applies to independent network fitting when identical observations use the same fitting procedure and random schedule. It does not apply to a sequential encoder, whose earlier observations differ.

We verify data regeneration, wave-equation residuals and finite-difference convergence, exact terminal-observation equality and opposite velocities, split integrity, exact current-observation chance, and bit-identical independent encodings

Encoder selection uses reconstruction alone. The selected width-32, two-hidden-layer SIREN has $\omega=10$ and 2,241 parameters. Its validation frame MSE is $2.4\times10^{-4}$, compared with $1.0\times10^{-3}$ for the width-16 alternative. Training uses 200 Reptile outer steps and 100 full-grid Adam steps per frame at learning rate $3\times10^{-3}$, with fresh optimizer state and $\beta=0$. All nine priors are frozen before hidden-state labels or result data are opened.

Three data seeds are crossed with three encoder seeds, with 384 held-out physical pairs. Paired-bootstrap intervals resample the physical pairs after averaging encoder effects within pair; crossed data/encoder cells are not treated as independent observations.

For the main nine-cell probe experiment, standardization and PCA are fitted on support data, and probe ranks and regularization strengths are selected on validation data. Probe selection uses four CPU threads. The decoded-field control uses float32 PCA, whose result can depend on the BLAS thread count; the supplement records this setting and the numerical software versions.

\subsection{Endpoint probes and reconstruction leakage}
\begin{table}[htbp]
\caption{\textbf{Wave-memory results for each seed cell.} Sign accuracy is in percent. Current-field and independent-weight probes are exactly 50\% in every row. Velocity $R^2$ uses sequential weights; Pair RMSE measures the difference between paired sequential reconstructions.}
\label{tab:app-collision}
\centering
\small
\setlength{\tabcolsep}{4pt}
\renewcommand{\arraystretch}{1.05}
\begin{tabular}{rrrrrrr}
\toprule
Data seed & Enc.\ seed & Sequential & Seq.\ decoded & Residual & Velocity $R^2$ & Pair RMSE \\
\midrule
14011 & 101 & 67.2 & 56.6 & 58.2 & 0.229 & 0.0062 \\
14011 & 211 & 68.0 & 54.7 & 50.8 & 0.216 & 0.0066 \\
14011 & 307 & 68.0 & 52.0 & 58.6 & 0.217 & 0.0065 \\
25031 & 101 & 64.8 & 57.4 & 59.0 & 0.234 & 0.0064 \\
25031 & 211 & 67.6 & 57.0 & 54.7 & 0.222 & 0.0066 \\
25031 & 307 & 68.4 & 52.7 & 53.5 & 0.200 & 0.0065 \\
36061 & 101 & 72.7 & 56.6 & 59.0 & 0.235 & 0.0062 \\
36061 & 211 & 70.7 & 55.5 & 55.9 & 0.216 & 0.0065 \\
36061 & 307 & 69.9 & 57.8 & 57.8 & 0.211 & 0.0065 \\
\bottomrule
\end{tabular}
\end{table}

Current-observation and independent-weight sign probes score exactly $50.0\%$ in every cell. Sequential weights give mean accuracy $68.6\%$, minimum $64.8\%$, and seed standard deviation $2.3$. The mean gain is $18.6$ points, with paired bootstrap 95\% interval $[16.5,20.6]$. A ridge probe of the full terminal velocity gives $R^2=0.22$, while the current-observation controls give zero.

Sequential reconstructions are not exactly equal across the paired histories. Their decoded-field sign probes reach $55.6\%$ on average, and terminal-residual probes reach $56.4\%$. Median pairwise reconstruction root mean squared error is $0.00645$. These measurements distinguish two claims: the final observed field lacks the sign information, but the sequentially decoded field need not lack it. Above-chance prediction from weights establishes retained history, not an exclusively output-invisible memory.

\subsection{Follow-up: fitting budget and history length}
\label{app:wave-followup}

\paragraph{History length.} Endpoint sign accuracy is 50.0\%, 60.7\%, 55.7\%, 57.7\%, and 68.6\% for 1, 2, 4, 8, and 16 fitted observations. The 16-frame condition exceeds every shorter-history condition under paired intervals. The result supports accessible historical information at the endpoint but not monotonic accumulation with history length.

\paragraph{Fitting budget.} From 25 to 500 Adam steps per frame, endpoint-weight accuracy remains between 67.7 and 68.6 and velocity \(R^2\) remains between 0.215 and 0.222. Thus this range does not show useful-memory washout as fitting increases. Decoded-field and residual probes follow different budget trends, so the weight signal is not explained simply by the measured reconstruction leakage.

\section{Tangent diagnostics and retention tests}
\label{app:retention}

\subsection{Matched and mismatched local updates}
\label{app:tangent-protocol}
At a transition's current weights $\theta_t$, let the residual to the next observation be $r_t=u_{t+1}-F(\theta_t)$. The tangent diagnostic solves
\[
\delta_t^*=\arg\min_\delta\left[\|J_t\delta-r_t\|_2^2+\lambda\|\delta\|_2^2\right].
\]
The mismatched control uses another transition's Jacobian, with the same targets, ridge scale, and downstream scoring. On shear flow the ridge scale is selected using support fidelity and is $\lambda=0.1$. This changes the local update model, rather than selectively changing geometry inside the original nonlinear encoder.

\begin{table}[t]
\caption{\textbf{Replacing the Jacobian changes a local update model on \GS{}.} Accuracy (\%) over nine seed cells, using 128 Jacobian points per transition. Matched beats mismatched in all nine cells. This diagnostic run reports its own raw comparator.}
\label{tab:tangent}
\centering
\small
\setlength{\tabcolsep}{5pt}
\renewcommand{\arraystretch}{1.08}
\begin{tabular}{lccc}
\toprule
Representation / update & Seed mean & Ensemble & Seed SD \\
\midrule
Raw-state dynamics            & 57.0 & 53.3 & 3.8 \\
Independent nonlinear INR     & 59.0 & 55.6 & 4.0 \\
Sequential nonlinear INR      & \textbf{64.9} & \textbf{65.6} & \textbf{1.2} \\
\midrule
Matched tangent update        & 46.5 & 54.4 & 4.1 \\
Mismatched tangent update     & 30.1 & 42.2 & 6.3 \\
\bottomrule
\end{tabular}
\end{table}

On \GS{}, the primary diagnostic uses all 15 transitions and 128 fixed spatial points per transition. Hyperparameters are selected on validation initial conditions 100--119 and evaluated on held-out initial conditions 145--159. Matched updates have cosine $0.159$ with nonlinear displacements, compared with $0.011$ for mismatched updates. Correlation of pairwise displacement geometry is $0.146$ versus $0.027$. Classification improves by $16.4$ seed-mean points or $12.2$ ensemble points, positive in all nine cells.

Sampling resolution matters. At 32 spatial points, the seed-mean gain drops to $7.4$ points, with six of nine positive cells and an ensemble difference of $-4.4$ points. The tangent model remains an imperfect description of the nonlinear displacement even at the higher resolution.

The same intervention fails on historical shear flow. Matched ensemble accuracy is $26.8\%$ (seeds $25.9/30.4/28.6$), versus $28.6\%$ mismatched (seeds $24.1/33.0/32.1$) and $38.4\%$ nonlinear. This occurs despite greater local fidelity than on \GS{}: matched displacement cosine and geometry correlation are $0.537/0.679$, versus $0.074/0.076$ mismatched. The nonlinear comparator differs from Table~\ref{tab:app-re45} because this protocol selects rank and ridge across three validation encoders. The diagnostic supports a system-dependent role for tangent identity, not a general causal explanation of nonlinear memory.

\subsection{The retention rule and its conditions}
\label{app:retention-law}
Consider two runs of gradient descent on the same frozen local model and target, differing only by a small initial parameter perturbation $e_0$. For loss $\frac12\|J\delta-r\|_2^2$, their difference follows
\[
e_{s+1}=(I-\eta J^\top J)e_s.
\]
If $e_0$ lies in an eigenvector of $J^\top J$ with eigenvalue $\kappa_i$, its signed amplitude after $s$ steps is multiplied by $(1-\eta\kappa_i)^s$. This gives Equation~\ref{eq:retention}. Under $0\le\eta\kappa_i\le1$, larger eigenvalues correspond to faster decay without sign changes. This condition makes precise the easy-to-change, easy-to-overwrite interpretation.

For a fixed positive-definite preconditioner $P$, write $e=P^{1/2}\widetilde e$. The corresponding symmetric matrix is $P^{1/2}J^\top JP^{1/2}$, with the same nonzero eigenvalues as $JPJ^\top$. The same mode-wise rule applies in these coordinates. An alternative squared-loss normalization rescales the curvature or effective step size. A changing Jacobian, a changing preconditioner, or retained optimizer state is outside this exact frozen-model calculation.

Implementation checks validate the linear rule to maximum absolute error $1.33\times10^{-15}$ and the preconditioned version to relative error $2.51\times10^{-14}$. Median nonlinear locality errors are $5.3\times10^{-8}$, $1.6\times10^{-7}$, and $5.3\times10^{-7}$ at the small, medium, and large perturbation amplitudes tested.

\begin{table}[htbp]
\caption{\textbf{Predicted and measured retention under controlled updates.} Correlation compares mode-wise retention for each transition. A log-retention slope of one is exact agreement. Slow and fast columns give retention after 16 steps.}
\label{tab:app-retention}
\centering
\small
\setlength{\tabcolsep}{4pt}
\renewcommand{\arraystretch}{1.08}
\begin{tabular}{lrrrrr}
\toprule
System & Instances & Median Spearman & Median log slope & Slow ret.\ $s{=}16$ & Fast ret.\ $s{=}16$ \\
\midrule
\AC{}       & 216 & 0.986 & 1.005 & 0.998 & 0.259 \\
\GS{} & 324 & 0.945 & 0.992 & 0.999 & 0.579 \\
\bottomrule
\end{tabular}
\end{table}

At saved sequential endpoints, the controlled-update measurements agree closely with the retention rule (Table~\ref{tab:app-retention}). Across seed combinations, median correlations range from $0.982$ to $0.989$ for \AC{} and $0.910$ to $0.964$ for \GS{}. These controlled measurements do not make the formula exact for the practical Adam encoder.

\subsection{Fitting budget in the practical encoder}
\begin{table}[htbp]
\caption{\textbf{More fitting improves reconstruction but not always classification.} Validation results. Raw accuracy is fixed at 41.7\% for \AC{} and 61.1\% for \GS{}.}
\label{tab:app-budget}
\centering
\small
\setlength{\tabcolsep}{4pt}
\renewcommand{\arraystretch}{1.05}
\begin{tabular}{lrrrr}
\toprule
System & Steps/frame & Seq.\ MSE & Indep.\ acc. & Seq.\ acc. \\
\midrule
\AC{} & 1    & 0.695 & 26.7 & 25.7 \\
\AC{} & 50   & 0.059 & 37.2 & 64.6 \\
\AC{} & 200  & 0.043 & 42.7 & \textbf{69.4} \\
\AC{} & 1000 & \textbf{0.035} & 45.1 & 60.1 \\
\midrule
\GS{} & 1    & 1.128 & 52.2 & 56.6 \\
\GS{} & 50   & 0.679 & 55.5 & \textbf{69.2} \\
\GS{} & 100  & 0.599 & 52.0 & 68.8 \\
\GS{} & 1000 & \textbf{0.395} & 57.2 & 64.1 \\
\bottomrule
\end{tabular}
\end{table}

Reconstruction improves as the fitting budget grows, but classification is not monotone. \AC{} sequential accuracy peaks at $69.4\%$ with 200 steps per frame, then falls to $60.1\%$ with 1,000. \GS{} peaks at $69.2\%$ with 50 steps and falls to $64.1\%$ with 1,000. Independent fits also improve with more steps; nevertheless, sequential fits lead at every tested \AC{} budget from five steps onward and at every tested \GS{} budget.

These are validation curves. Their raw accuracies, $41.7\%$ and $61.1\%$, should not be substituted for held-out comparators elsewhere. 

\subsection{Selective erasure does not identify slow modes as the memory carrier}
\label{app:erasure}
The erasure test projects the measured sequential-minus-independent history component onto either the slowest or fastest quartile of non-null instantaneous tangent modes. It removes that component with strengths chosen to match the resulting change in decoded outputs. The classifier is held fixed.

Mean slow-minus-fast accuracy damage is 1.04 points. Slow-mode erasure is more damaging in three of nine cells, with five ties and one opposite ordering. The decoded-distortion ratio is 1.001, indicating that the output-matching control succeeds. The result therefore does not identify the frozen slow quartile as the useful-memory carrier, and it does not invalidate the retention rule.

\subsection{Differentiating through multiple Adam fits: numerical limits}
\label{app:fh}
The finite-horizon analysis differentiates through the actual fresh-state Adam fitting blocks, composing the derivatives across frames. Its goal is to find endpoint directions most sensitive to observations several frames earlier, rather than using only the current tangent kernel.

One-frame replay relative parameter error is $1.9\times10^{-7}$ on \AC{} and $3.6\times10^{-8}$ on \GS{}, and the Jacobian-vector/vector-Jacobian adjoint check is $1.0\times10^{-14}$, supporting the implementation at one frame. At eight frames, replay error rises to \(4.69\times10^{-5}\) and the estimated leading singular value shows a drift of 0.264. We therefore do not perform the intended causal erasure intervention because the composed derivative is not numerically stable enough for that test. This leaves the causal question unanswered; it is not evidence that finite-horizon memory directions do not exist.

\section{Controlled loops: implementation, diagnostics, and boundaries}
\label{app:holonomy-protocol}

\subsection{Exact transport and numerical eligibility}
The construction uses complete \(12\times12\) periodic output grids and a direct QR solve, avoiding normal equations.
\[
R^\top v=\dot y,\qquad \dot\theta=Qv.
\]
This is the minimum-norm parameter velocity when $J$ has full row rank. It uses no damping, singular-value truncation, or normal equations. An untruncated reduced singular-value decomposition independently checks the same parameter velocity.

The experiment uses three base fields crossed with three initializations, fixed loop-direction pairs, and scales $\{0.005,0.01,0.02,0.04\}$. Fourth-order Runge--Kutta integration is run at coarse and fine resolutions in double precision on a single CPU thread. The parameter metric is the identity. ``Exact'' refers to the transport rule and full sampled-output constraint; numerical integration accuracy is separately measured.

All nine base-field/initialization combinations satisfy the prerequisites: rank $144/144$, maximum Jacobian condition number $1.53\times10^5$, base QR residual at most $6.6\times10^{-16}$, QR/SVD parameter-velocity disagreement at most $4.1\times10^{-12}$, and transport solve residual at most $1.1\times10^{-15}$.

\subsection{Full loop diagnostics and reproduction}
\begin{table}[t]
\caption{\textbf{Full controlled-loop diagnostics.} Local medians and slopes use nine base-field/initialization combinations at the two smallest loop scales, 0.005 and 0.01 (18 combination/scale records); figure-eight suppression uses the available control records at these scales. The minimum measured tangent-kernel change covers all nine combinations and all four scales.}
\label{tab:g1c}
\centering
\small
\setlength{\tabcolsep}{5pt}
\renewcommand{\arraystretch}{1.08}
\begin{tabular}{llr}
\toprule
Diagnostic & Ideal & Result \\
\midrule
Median complete-field relative closure (fine RK4) & $0$ & $2.99\times10^{-14}$ \\
Closure reduction, coarse$\to$fine integration & large & $16.5\times$ \\
Net parameter-change local slope in loop scale & $2$ & $2.023$ \\
Tangent-kernel-change local slope & $2$ & $1.979$ \\
Forward/reverse parameter cosine & $-1$ & $-0.999954$ \\
Direct Lie-bracket cosine & $+1$ & $0.9948$ \\
Kernel-response prediction cosine & $+1$ & $0.9965$ \\
Tangent-score response Spearman & $+1$ & $0.9903$ \\
Equal-area figure-eight suppression & large & $7.4\times10^{9}\times$ \\
Minimum tangent-kernel change (all combinations/scales) & $>0$ & $0.354$ \\
Constant-Jacobian control, kernel change & $0$ & exactly $0$ \\
\bottomrule
\end{tabular}
\end{table}

The output-loop error decreases with finer integration while nonzero parameter and kernel changes remain. Across all 36 loops at the four scales, the median fine-integration relative closure is $1.71\times10^{-12}$ and the maximum is $4.64\times10^{-9}$; the smaller value in Table~\ref{tab:g1c} summarizes the two smallest scales. The remaining diagnostics in Table~\ref{tab:g1c} show the expected area scaling, orientation reversal, figure-eight suppression, and agreement with local Lie-bracket, kernel-response, and tangent-score predictions. The constant-Jacobian decoder has exactly zero kernel change. Together, these tests support curvature of the nonlinear parameterization as the source of the controlled effect. They establish equality of all 144 sampled outputs, not equality of the continuous function away from those coordinates.

\subsection{Comparison with ordinary fitting}
The optimizer comparison fits the same loop targets with SGD and Adam variants. Stateless SGD uses learning rate $0.009$ and 100 steps per target, selected on separate validation base/initialization combinations for one-way target tracking. The evaluation uses 32 targets per loop side on six held-out base/initialization combinations at all four scales.

Practical and exact kernel-change norms correlate (0.9148), but their directions do not: the median cosine is 0.0247. Forward/reverse kernel-change cosines remain positive for SGD (+0.6676), fresh Adam (+0.9071), and carried Adam (+0.9654), so the orientation reversal characteristic of the exact transport is absent. Increasing target density from 4 to 32 reduces tracking MSE from \(1.6\times10^{-6}\) to \(2.7\times10^{-7}\), but changes the exact-kernel cosine only from -0.222 to +0.105.

A zero-loop control repeatedly fits the unchanged starting field for 128 fresh blocks per base/initialization combination and produces exactly zero loss, weight change, and kernel change. Ordinary fitting therefore fails the characteristic exact-transport signatures in this tested setting. This is the relevant boundary on using the construction to explain the practical sequential encoder.

The controlled-loop experiment therefore establishes output-preserving changes in tangent geometry, but it does not show that ordinary SGD/Adam reproduces the transport or that such geometry edits improve subsequent learning.

\end{document}